\documentclass[runningheads]{llncs}
\usepackage[T1]{fontenc}
\usepackage{graphicx}
\usepackage{amsmath,amssymb,amsfonts}
\usepackage{algorithm}
\usepackage{algpseudocode}
\usepackage{microtype}
\usepackage{cite}
\usepackage[section]{placeins}
\begin{document}
\title{Bayesian Decision Trees Inspired from Evolutionary Algorithms}
\author{Efthyvoulos Drousiotis\inst{1} \and
Alexander M, Phillips\inst{1}\and
Paul G, Spirakis\inst{2}\and
Simon Maskell\inst{1}}
\authorrunning{E Drousiotis et al.}

\institute{Department of Electrical Engineering and Electronics, University of Liverpool, Liverpool L69 3GJ, UK;
\email{\{E.Drousiotis, A.M.Philips, S.Maskell\}@liverpool.ac.uk} \and
Department of Computer Science, University of Liverpool, Liverpool L69 3BX, UK\\
\email{P.Spirakis@liverpool.ac.uk}\\
}
\maketitle              % typeset the header of the contribution
\sloppy
\begin{abstract}
Bayesian Decision Trees (DTs) are generally considered a more advanced and accurate model than a regular Decision Tree (DT) because they can handle complex and uncertain data. Existing work on Bayesian DTs uses Markov Chain Monte Carlo (MCMC) with an accept-reject mechanism and sample using naive proposals to proceed to the next iteration, which can be slow because of the burn-in time needed. We can reduce the burn-in period by proposing a more sophisticated way of sampling or by designing a different numerical Bayesian approach. In this paper, we propose a replacement of the MCMC with an inherently parallel algorithm, the Sequential Monte Carlo (SMC), and a more effective sampling strategy inspired by the Evolutionary Algorithms (EA). Experiments show that SMC combined with the EA can produce more accurate results compared to MCMC in 100 times fewer iterations.


\end{abstract}
\section{Introduction and Relevant Work}\label{relevantwork}

In Bayesian statistics, obtaining and calculating random samples from a probability distribution is challenging. Markov Chain Monte Carlo (MCMC) is a widely used method to tackle this issue. MCMC can characterise a distribution without knowing its analytic form by using a series of samples. MCMC has been used to solve problems in various domains, including biology\cite{valderrama2019mcmc}, forensics\cite{taylor2014interpreting}, education\cite{drousiotis2021early, drousiotis2021capturing}, and chemistry\cite{dumont2021quantification}, among other areas. Monte Carlo applications are generally considered embarrassingly parallel since each chain can run independently on two or more independent machines or cores.
Nevertheless, the principal problem is that processing within each chain is not embarrassingly parallel. When the feature space and the proposal are computationally expensive, we can only do a little to improve the running time. As mentioned before, running multiple independent MCMC chains in parallel can be a strategy for parallelization, but it is not very effective. Even though you increase the number of chains and decrease the number of samples produced from each chain, the burn-in process for each chain would remain unchanged.

Much research has been done to improve the efficiency of MCMC methods. The improvements can be grouped into several categories\cite{robert2018accelerating}. One of these categories is based on understanding the geometry of the target density function. An example of this is Hamiltonian Monte Carlo\cite{duane1987hybrid} (HMC), which uses an auxiliary variable called momentum and updates it using the gradient of the density function. Different methods, such as symplectic integrators of various precision levels, have been developed to approximate the Hamiltonian equation\cite{blanes2014numerical}. HMC tends to generate less correlated samples than the Metropolis-Hastings algorithm, but it requires the gradient of the density function to be available and computationally feasible.

Another approach involves dividing complex problems into smaller and more manageable components. For example, as discussed earlier, using multiple parallel MCMC chains to explore the parameter space and then combining the samples obtained from these chains\cite{mykland1995regeneration}. However, this approach does not achieve faster convergence of the chains to the stationary distribution. This is because all chains have to converge independently of each other in MCMC. It is also possible to partition the data space, which is already implemented in the context of Bayesian DTs\cite{pakdd}, or partition the parameter space \cite{pmlr-v51-basse16a}into simpler pieces that can process independently which are proven not to be much effective. 

MCMC DTs simulations can take a long to converge on a good model when the state space is large and complex. This is due to both the number of iterations needed and the complexity of the calculations performed in each iteration, such as finding the best features and structure of a DT. MCMC, in general, and as will be explained detailed in section~\ref{Section3}, generates samples from a specific distribution by proposing new samples and deciding whether to accept them based on the evaluation of the posterior distribution. Because the current sample determines the next step of the MCMC, it can be not easy to process a single MCMC chain simultaneously using multiple processing elements. A method is \cite{drousiotis2023single} proposed to parallelise  a single chain on MCMC Decision trees. However, the speedup is not always guaranteed.

Another method of speeding up MCMC, focuses on enhancing the proposal function, which is the approach we pursue in this paper. This can be achieved through techniques such as simulated tempering\cite{marinari1992simulated}, adaptive MCMC\cite{douc2007convergence}, or multi-proposal MCMC\cite{liu2000multiple}.
Having a good proposal function in Markov Chain Monte Carlo (MCMC) and, in general, Monte Carlo methods is crucial for the efficiency and accuracy of the algorithm. Poor proposals can lead to slow convergence, poor mixing, and biased estimates. Good proposal functions can efficiently explore the target distribution and reduce the correlation between successive samples. 

Several papers describe novelties specifically on Bayesian DTs, focusing on different improvements. For example, \cite{pakdd, drousiotis2023single, taddy2011dynamic}, contributed towards the runtime enhancement, \cite{wu2007bayesian} explored a new novel proposal move called "radical restructure," which changes the structure of the tree ($T$), without changing the number of leaves or the partition of observations into leaves. Moreover, \cite{chipman1998bayesian} proposed different criteria for accepting or rejecting the $T$ on MCMC, such as posterior probability, marginal likelihood, residual sum of squares, and misclassification rates. The general approach to improving the proposal function in Monte Carlo methods has been introduced previously. However, in the context of Bayesian DTs, there is still enough space for exploration.
 
Our contribution is then:

\begin{itemize}
  \item We describe for the first time a novel algorithm inspired by the Evolutionary Algorithms to improve the proposal function of the Bayesian DTs that uses an inherently parallel algorithm, a Sequential Monte Carlo (SMC) sampler, to generate samples.
\end{itemize}

\section{Bayesian Decision Trees}\label{Section3}

\begin{figure}
    \centering
    \includegraphics[width=0.3\textwidth]{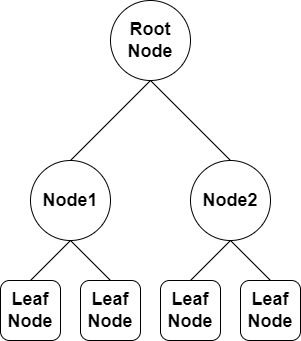}
    \caption{Decision Tree}
    \label{DT}
\end{figure}

A DT operates by descending a tree $T$. The process of outputting a classification probability for a given datum starts at a root node (see Figure \ref{DT}). At each non-leaf node, a decision as to which child node to progress to is made based on the datum and the parameters of the node. This process continues until a leaf node is reached. At the leaf node, a node-specific and datum-independent classification output is generated.

Our model describes the conditional distribution of a value for $Y$ given the corresponding values for $x$, where $x$ is a vector of predictors %$[x = (x_1,x_2,...,x_p)]$, 
and $Y$ the corresponding values that we predict. We define the tree to be $T$ such that the function of the non-leaf nodes is to (implicitly) identify a region, $A$, of values for $x$ for which $p\left(Y|x\in A\right)$ can be approximated as independent of the specific value of $x\in A$, ie as $p(Y|x)\approx p(Y|\phi_j,x\in A)$. This model is called a probabilistic classification tree, according to the quantitative response $Y$.

For a given tree, $T$, we define its depth to be $d(T)$, the set of  leaf nodes to be $L(T)$ and the set of non-leaf nodes to be $\bar{L}(T)$. The $T$ is then parameterised by: the set of features for all non-leaf nodes, $k_{\bar{L}(T)}$;  the vector of corresponding thresholds, $c_{\bar{L}(T)}$; the parameters, $\phi_{L(T)}$, of the conditional probabilities associated with the leaf nodes. This is such that the parameters of the $T$ are $\theta(T)=[k_{\bar{L}(T)},c_{\bar{L}(T)},\phi_{L(T)}]$ and $\theta(T)_j$ are the parameters associated with the $j$th node of the $T$, where:
\begin{align}
\theta(T)_j = \left\{\begin{array}{ll}
\left[k_j,c_j\right] &j\in \bar{L}(T)\\
\phi_j& j\in L(T).\end{array}\right.
\end{align}

Given a dataset comprising $N$ data, $Y_{1:N}$ and corresponding features, $x_{1:N}$, and since DTs are specified by $T$ and $\theta(T)$, a Bayesian analysis of the problem proceeds by specifying
a prior probability distribution, $p(\theta(T) ,T)$ and associated likelihood, $p(Y_{1:N}|T,\theta(T),x_{1:N})$. Because $\theta(T)$ defines the parametric model for $T$, it will usually be convenient to adopt the following structure for the joint probability distribution of $N$ data, $Y_{1:N}$, and the $N$ corresponding vectors of predictors, $x_{1:N}$: 
\begin{align}\label{fullformula}
 p(Y_{1:N},T,\theta(T)|x_{1:N}) =& p(Y_{1:N}|T,\theta(T),x_{1:N})p(\theta(T),T)\\
 =&p(Y_{1:N}|T,\theta(T),x_{1:N})p(\theta(T)|T)p(T)
\end{align}
which we note is proportional to the posterior, $p(T,\theta(T)|Y_{1:N},x_{1:N})$, and where we assume
\begin{align}
     p(Y_{1:N}|T,\theta(T),x_{1:N}) &= \prod_{i = 1}^{N} p(Y_i|x_i,T,\theta(T))   \label{labels probabiliteis}\\
    p(\theta(T)|T) &= \prod_{j\in T}p(\theta(T)_j|T) \\&= \prod_{j\in T} p(k_j|T)p(c_j|k_j,T)\label{features and thresholds}\\
    p(T) &= \frac{a}{(1+d(T))^\beta}\label{prior}
\end{align}

Equation~\ref{labels probabiliteis} describes the product of the probabilities of every data point, $Y_i$, being classified correctly given the datum's features, $x_i$, the $T$ structure, and the features/thresholds, $\theta(T)$, associated with each node of the $T$.
At the $j$th node, equation~\ref{features and thresholds} describes the product of possibilities of picking the $k_j$th feature and corresponding threshold, $c_j$, given the $T$ structure.
Equation~\ref{prior} is used as the prior for the $T$. This prior is recommended by \cite{chipman2010bart} and three parameters specify this prior: the depth of the $T$, $d(T)$; the parameter, $a$, which acts as a normalisation constant; the parameter, ${\beta > 0}$, which specifies how many leaf nodes are probable, with larger values of $\beta$ reducing the expected number of leaf nodes. $\beta$ is crucial as this is the penalizing feature of our probabilistic $T$ which prevents an algorithm that uses this prior from over-fitting and allows convergence to occur\cite{rovckova2019theory}. Changing $\beta$ allows us to change the prior probability associated with ``bushy'' trees, those whose leaf nodes do not vary too much in depth.

An exhaustive evaluation of equation~\ref{fullformula} over all trees will not be feasible, except in trivially small problems, because of the sheer number of possible trees. 

Despite these limitations, Bayesian algorithms can still be used to explore the posterior. Such algorithms simulate a chain sequence of trees, such as:

\begin{equation}\label{chain}
    T_0, T_1, T_2,....,T_n
\end{equation}
which converge in distribution to the posterior, which is itself proportional to the joint distribution,
$p(Y_{1:N}|T,\theta(T),x_{1:N})p(\theta(T)|T)p(T)$, specified in equation~\ref{fullformula}. We choose to have a simulated sequence that gravitates toward regions of the higher posterior probability. Such a simulation can be used to search for high-posterior probability trees stochastically. We now describe the details of  algorithms that achieve this and how they can be implemented.

\subsection{Stochastic Processes on Trees}\label{moves}

To design algorithms that can search the space of trees stochastically, we first need to define a stochastic process for moving between trees. More specifically, we consider the following four kinds of move from one $T$ to another:
\begin{itemize}
\item Grow(G) : we sample one of the leaf nodes, $j\in L(T)$, and replace it with a new node with parameters, $k_j$ and a $c_j$, which we sample uniformly from their parameter ranges.
\item Prune(P) : we sample the $j$th node (uniformly) and make it a leaf.
\item Change(C) : we sample the $j$th node (uniformly) from the non-leaf nodes, $\bar{L}(T)$, and sample $k_j$ and a $c_j$ uniformly from their parameter ranges.
\item Swap(S) : we sample the $j_1$th and $j_2$th nodes uniformly, where $j_1\neq j_2$, and swap $k_{j_1}$ with $k_{j_2}$ and $c_{j_1}$ with $c_{j_2}$.
\end{itemize}

We note that there will be situations (eg when moving from a $T$ with a single node) when some moves cannot occur. We assume each `valid' move is equally likely and this then makes it possible to compute the probability of transition from one $T$, to another, $T'$, which we denote $q\left(T',\theta(T')|T,\theta(T)\right)$.

\section{Our Approach on Evolutionary Algorithms}

Evolutionary Algorithms (EA) mimic living organisms' behavior, using natural mechanisms to solve problems\cite{evolutionary_algo}. In our approach, the optimisation problem is represented as a multi-dimensional space on which the population lives (in our case, the population is the total number of trees). Each location on the solution space corresponds to a feasible solution to the optimisation problem. The optimal solution is found by identifying the optimal location in the solution space.

Pheromones play a crucial role in evolutionary algorithms as they are used  for communication among the population \cite{pheromones1, pheromones2}. When a member of the population moves, it receives pheromones from the other member of the population and uses them to determine its next move. Once all members of the population reach new locations, they release pheromones; the concentration and type of pheromones released depend on the objective function or fitness value at that location. The solution space is the medium for transmitting pheromones, allowing individuals to receive and be affected by the pheromones released by other individuals, creating a global information-sharing model.

The population will gradually gain a rough understanding of global information through their movements, which can significantly benefit the optimisation process. In our approach, the EA can use the solution space as a memory to record the best and worst solutions produced in each iteration. Once the positioning stage is finished, all pheromones on the solution space are cleared. To guide the optimisation process, the most representative extreme solutions are selected from the recorded solutions, and the corresponding locations are updated with permanent pheromones. Unlike permanent pheromones, temporary pheromones only affect the movements of trees in the next iteration.

\section{Methods}
\subsection{Conventional MCMC}\label{sec:MCMC}

One approach is to use a conventional application of Markov Chain Monte Carlo to DTs, as found in~\cite{drousiotis2023single}.

More specifically, we begin with a tree, $T_0$ and then at the $i$th iteration, we propose a new $T'$ by sampling $T'\sim q\left(T',\theta(T')|T_i,\theta(T_i)\right)$. We then accept the proposed $T'$ by drawing $u\sim U([0,1])$ such that:
\begin{align}
T_{i+1} = \left\{\begin{array}{ll} T' & u\leq\alpha(T'|T)\\
T_i & u > \alpha(T'|T)\end{array}\right.
\end{align}
where we define the acceptance ratio, $\alpha(T',T)$ as:
\begin{align}\label{alpha}
\alpha(T'|T) = \frac{p(Y_{1:N}|T,\theta(T),x_{1:N})}{p(Y_{1:N}|T',\theta(T'),x_{1:N})}\frac{q\left(T,\theta(T)|T',\theta(T')\right)}{q\left(T',\theta(T')|T,\theta(T)\right)}
\end{align}.

This process proceeds for $n$ iterations. We take the opportunity to highlight that this process is inherently sequential in its nature.

\subsection{Evolutionary Algorithm in Bayesian Decision Trees}\label{hybrid_algo} 

\subsubsection{Initializing population}\label{initialise}
The initial population plays a crucial role in the solutions' quality. An initial population with a good mix of diversity and a substantial number of trees can help improve the optimisation performance of the algorithm. To create a diverse initial population, a random method is often employed. This approach helps to ensure that the algorithm can perform a global search efficiently, is easy to use, and has a diverse set of individuals in the initial population.

The population size of trees is invariable, denoted as $n$. The location of every $T_i$ in the $D$ - dimensional space can be described as $: T=(T_1^1,T_2^2,…,T_n^d,…,T_N^D)$. According to the value of our objective function $p(Y1:N , T, \theta(T 
)|x1:N )$ , the fitness value of location $d$ and $T_i$ can be measured. By comparing the current location of each $T_i$ in the initial population, the optimal location and the worst location in the initial population were obtained, and the value of the objective function of the optimal location in the initial population was recorded.

\subsubsection{Positioning stage}

\begin{figure}
    \centering
    \includegraphics[width=0.5\textwidth]{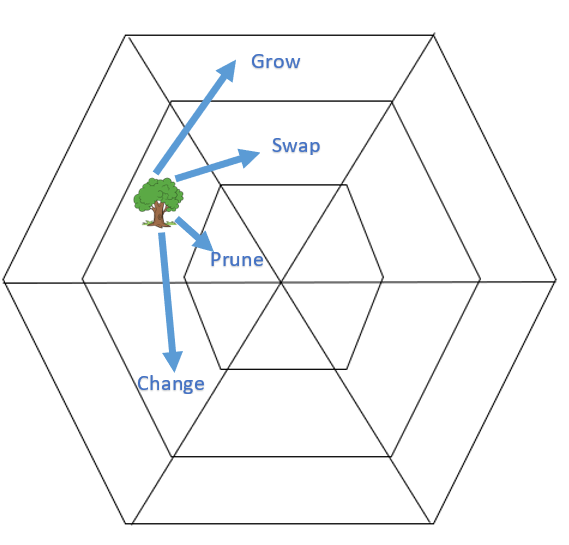}
    \caption{Positioning Stage}
    \label{Positioning}
\end{figure}

In the positioning stage, in our use case, trees release permanent and temporary pheromones. The solution space records the locations of the terrible solution and the excellent solution produced by each iteration. While all trees move to the new location, the pheromones are updated differently, which will be discussed in section~\ref{agg}. The process in the positioning stage is shown in Figure~\ref{Positioning}
In our case, the possible movements of the $T_i$ are those described in section~\ref{moves}. When the proposed move is the $Grow$, we search for possible solutions in a higher dimensional space; when the proposed move is the $Prune$, we explore for possible solutions in lower dimensional space; and when the proposed moves are $Change$ and $Swap$, we search for solutions on the same dimensional space.

\subsubsection{Permanent Pheromones}\label{marked}
$Permanent\; pheromones$ have persistent effects. On each iteration, we evaluate each $T_i$ to compute the value $\alpha_n$ (see equation \ref{alpha}). If $\alpha_n$ is greater than a uniform number between $[0,1]$, we store the $T_i$ and the stochastic move associated with $positive\; exploration$ and $effective\; moves$, respectively. If $\alpha_n$ is less than a uniform number between $[0,1]$ we store the $T_i$ and the stochastic move associated, on $negative \;explartion$ and $innefective\; moves$ respectively($negative\; explartion$ and $innefective\; moves$ will be used on $temporary\; pheromones$).
We repeat the procedure above for all trees. We then calculate the $permanent\; pheromones$ given the $effective\; moves$. $Permanent\; pheromones$ is a single list with $4$ real numbers, representing the possibilities of each one of the $4$ stochastic moves to be selected on the next iteration. We update the $permanent \; pheromones$ by adding to each possibility on the $permanent\; pheromones$, the number of times each move is in the $effective\; moves$ list, and we then divide each element of the list by the sum of the $effective \;moves$ (we normalise to add up to 1). We have also designed a mechanism to avoid having a biased $permanent\; pheromones$ list. For example, in the first stages of the algorithm, it is common the $Grow$ move to be a more useful proposal compared to the $Prune$. In such cases, if a specific move has more than $80\%$ possibilities to be chosen, we re-weight the list by setting the dominant move having  $40\%$ possibilities to be selected and the rest having $20\%$ possibilities to be chosen on the next iteration.  

\subsubsection{Temporary Pheromones}\label{agg}
$Temporary\; pheromones$ can only affect the movements of trees in the next iteration. We discussed above how we end up with a list called $effective\; moves$ and a list called $ineffective\; moves$. When the iteration ends,$temporary\; pheromones$ clear, and each $T_i$ will release new $temporary\; pheromones$, which will be recorded. Algorithm~\ref{pheromonesalgo} shows how we produce and store pheromones.

\subsection{Sequential Monte Carlo with EA}

We are considering a Sequential Monte Carlo(SMC) sampler\cite{del2006sequential} to handle the problem of sampling the trees from the posterior. After we have collected all the useful information from the movement of the trees, we now need to sample using the pheromones produced. As the algorithm~\ref{spcalgo} shows, there are three possible sampling techniques where one has a subcategory. We choose the sampling technique by drawing a uniform number between $[0,1]$. At this point, we need to specify that on each iteration, we draw a new uniform number, so each $T_i$ has the possibility to sample with a different strategy. 

The first sampling technique uses the $temporary\; pheromones$, and it has a possibility of $45\%$ to be chosen. As mentioned earlier, $temporary\; pheromones$ are produced during the previous iteration. This sampling technique has a subcategory, where the samples classified in $positive \;exploration$ use a different sampling technique from those listed on $negative\; exploration$. 
If the $T_i$ is listed in  $positive \;exploration$, we pick a stochastic move $m$ from the list with the $ineffective\; moves$ uniformly. We then remove all the identical $m$ from the $ineffective\; moves$ and sample uniformly with equal probabilities from the remaining $ineffective\; moves$.
If the $T_i$ is in $negative\; exploration$, we pick uniformly a stochastic move $m$ from the $effective\; moves$ list to sample the particular $T_i$.

%add examples here 

The second sampling technique uses the $permanent \; pheromones$, and it has a possibility of $45\%$ to be chosen. As mentioned earlier, $permanent \; pheromones$ are updated dynamically on every iteration, considering all the previous iterations. We sample the $T_i$ using the possibilities in the list we describe in subsection A$permanent \; pheromones$ in section~\ref{hybrid_algo}. 

The last sampling technique is straightforward. We use a list with stochastic moves, where each move has a uniform probability of being chosen. This sampling technique is unaffected by the pheromones. The main reason for including this technique is to ensure our algorithm is not biased, as this is the most common way of sampling in Bayesian DTs.

\begin{algorithm}
\caption{Pheromones Production Stage}\label{pheromonesalgo}
\begin{algorithmic}

\State Initialise $n$ trees($T$)
\State Sample trees $[T_0, T_1, ..., T_n]$
\State Initialise initial possibilities = $[p(G),p(P),p(C),p(S)] = [0.25,0.25,0.25,0.25]$
\State Initialise permanent pheromones = $[p(G),p(P),p(C),p(S)]$  \Comment{permanent pheromones}
\State Initialise $positive\;exploration $ list
\State Initialise $effective\;moves $ list \Comment{Temporary Pheromones}
\State Initialise $negative\;exploration$ list
\State Initialise $ineffective\;moves$ list \Comment{Permanent Pheromones}
\State iterations = $10$

\For{$ (i \leq iterations, i++ ) $}
    \State Evaluate trees $[T_0, T_1, ..., T_n]$ \State Store their acceptance probability $[\alpha_0, \alpha_1,...,\alpha_n]$
    \For{$ (s \leq n, s++ ) $}
        \State Draw a uniform number $u_1 \sim u[0,1]$
        \If{$\alpha_s > u_1$}
            \State append $T_s$ to list $positive\; exploration$
            \State append $T_s move$ to list $effective\; moves$
        \Else
            \State append $T_s$ to list $negative\;expolation$
            \State append $T_s move$ to list $ineffective\; moves$ 
        \EndIf
    \EndFor

    \State update $permanent\; pheromones$ list given the  $effective\; moves$ list

\EndFor
\end{algorithmic}
\end{algorithm}

\begin{algorithm}
\caption{SMC with EA}\label{spcalgo}
\begin{algorithmic}

    \For{$ i \leq n, i++$}
        \State Draw a uniform number $u_2 \sim u[0,1]$
        \If{$u_2 \leq 0.45$}
            \If{$T_i \;in\; positive\; exploration$}
                \State pick uniformly a move $m$ from $ineffective\;moves$
                \State Remove every identical $m$ from the $ineffective\;moves$
                \State Sample $T_i$ uniformly given the updated $ineffective\;moves$
            \EndIf
            \If{$T_i \;in\; negative\; exploration$} 
                \State Pick uniformly a move $m$ from $effective\; moves$
                \State Sample $T_i$ by applying the selected move $m$
            
            \EndIf
        \ElsIf{$u_2 > 0.45$ and $u_2 \leq 0.9$}
            \State Sample $T_i$ using $permanent\;pheromones$ 
        \Else 
            \State Sample $T_i$ using $initial\;possibilities$
        \EndIf
        
    \EndFor

    \State Empty $positive\;exploration$
    \State Empty $effective\;moves$ \Comment{Temporary Pheromones}
    \State Empty $negative\;exploration $
    \State Empty $ineffective\;moves$ \Comment{Permanent Pheromones}
    
\end{algorithmic}
\end{algorithm}

\section{Experimental Setup and Results}\label{Results}
To demonstrate the accuracy improvement we achieved through our proposed methods, we experiment on three publicly available datasets\footnote{https://archive.ics.uci.edu/ml/index.php} listed in Table~\ref{Datasets}. We acknowledge that the size of the datasets we conduct the experiments is small. The main reason is to have a fair comparison between the MCMC algorithm and SMC, as experiments show\cite{pakdd} that the former struggles to converge on an adequate time on big datasets, compared to the latter. We also aim to show that SMC, an inherently parallel algorithm combined with EA, can be a great fit within the context of big data. For each dataset, we have multiple testing hypotheses. Firstly, we compare the SMC-EA with MCMC on $1000$ iterations and $10$ chains for MCMC and $10$ $T$ for the SMC-EA, $100$ iterations and $100$ chains for MCMC and $100$ $T$ for the SMC-EA, and $10$ iterations with $1000$ chains for MCMC and $1000$ $T$ for the SMC-EA.

\begin{table}[]\caption{Datasets description}
\centering
\begin{tabular}{|c|c|c|}
\hline
\multicolumn{1}{|l|}{\textbf{Dataset}} & \multicolumn{1}{l|}{\textbf{Attributes}} & \multicolumn{1}{l|}{\textbf{Instances}} \\ \hline
Heart Disease & 75 & 303 \\ \hline
Lung Cancer & 56 & 32 \\ \hline
SCADI & 206 & 70 \\ \hline
\end{tabular}\label{Datasets}
\end{table}

This section presents the experimental results obtained using the proposed methods with a focus on accuracy improvement and the ability of the SMC-EA to evolve smoothly in a very short period of iterations. We obtained the following results using a local HPC platform comprising twin SP 6138 processors (20 cores, each running at 2GHz) with 384GB memory RAM. 
We use the same hyper-parameters $\alpha$ and $\beta$ for every contesting algorithm for testing purposes and a fair comparison and evaluation.

We tested both MCMC and SMC-EA with a 5-Fold Cross-Validation. Results indicate what is discussed in Section~\ref{initialise}. When we initialise more trees, in our case trees, we introduce a more diverse set, which helps improve the algorithm's optimization performance. More specifically, SMC-EA with $1000$ $T$ and $10$ iterations running SCADI dataset has an accuracy improvement of $\sim2\%$  and $\sim6\%$ compared to having $100$ $T$ with $100$ iterations, and $10$ $T$ with $1000$ iterations respectively. 
On the Heart Disease dataset SMC-EA with $1000$ $T$ and $10$ iterations has an accuracy improvement of $\sim3\%$  and $\sim4\%$ compared to having $100$ $T$ with 100 iterations and $10$ $T$ with $1000$ iterations respectively. 
On the Lung Cancer dataset SMC-EA with $1000$ $T$ and $10$ iterations, has an accuracy improvement of $\sim8\%$  and $\sim14\%$ compared to having $100$ $T$ with $100$ iterations, and $10$ $T$ with $1000$ iterations respectively. 

On the other hand, MCMC performs poorly when we have fewer iterations and more chains compared to SMC-EA. This is expected as MCMC needs adequate time to converge. When MCMC ran for more iterations, the algorithm made better predictions, and the accuracy achieved cannot be accepted as an acceptable threshold. SMC-EA has a $\sim12\%$ better predictive accuracy on the SCADI dataset compared to MCMC, $\sim7\%$ on Heart Disease, and $\sim17\%$ on Lung Cancer(see Tables~\ref{SCADI}, ~\ref{Heart}~and~\ref{Lung}). 

On the SMC-EA algorithm, trees do not have a leading $T_i$, and their movements are guided by interactions between them rather than one individual $T$ dominating the group of trees. This helps to prevent individualism and stagnation in the population's evolution. Furthermore, a diverse population of trees is more effectively handled by the approach we suggest, as we avoid falling into local optimisation. 
The stochastic nature of the SMC-EA algorithm helps in escaping local optimisation and achieving global optimisation. Combining positive and negative feedback from the different pheromones can incorporate the benefits of successful solutions while mitigating the negative effects of poor solutions. SMC-EA algorithm fully uses all the information on the solution space, avoiding unnecessary waste or duplication of information. 
The EA algorithm updates the positions of all trees by using a combination of their current and past positions within the population. This allows the algorithm to maintain a history of information, preventing rapid jumps and leading to a smooth algorithm evolution.

\begin{table}[H]\caption{SCADI dataset}
\centering
\begin{tabular}{|l|c|c|c|}
\hline
\textbf{Chains\_Trees} & \multicolumn{1}{l|}{\textbf{Iterations}} & \multicolumn{1}{l|}{\textbf{MCMC}} & \multicolumn{1}{l|}{\textbf{SMC-EA}} \\ \hline
10 & 1000 & 85 & 90.2 \\ \hline
100 & 100 & 63 & 94.2 \\ \hline
1000 & 10 & 57 & 96.6 \\ \hline
\end{tabular}\label{SCADI}
\end{table}

\begin{table}[H]\caption{Heart Disease dataset}
\centering
\begin{tabular}{|l|c|c|c|}
\hline
\textbf{Chains\_Trees} & \multicolumn{1}{l|}{\textbf{Iterations}} & \multicolumn{1}{l|}{\textbf{MCMC}} & \multicolumn{1}{l|}{\textbf{SMC-EA}} \\ \hline
10 & 1000 & 75.7 & 78.9 \\ \hline
100 & 100 & 73.5 & 79.2 \\ \hline
1000 & 10 & 67.7 & 82.7 \\ \hline
\end{tabular}\label{Heart}
\end{table}

\begin{table}[H]\caption{Lung Cancer dataset}
\centering
\begin{tabular}{|l|c|c|c|}
\hline
\textbf{Chains\_Trees} & \multicolumn{1}{l|}{\textbf{Iterations}} & \multicolumn{1}{l|}{\textbf{MCMC}} & \multicolumn{1}{l|}{\textbf{SMC-EA}} \\ \hline
10 & 1000 & 70.1 & 73.9 \\ \hline
100 & 100 & 69.7 & 79.1 \\ \hline
1000 & 10 & 69.1 & 87.2 \\ \hline
\end{tabular}\label{Lung}
\end{table}

Due to the small size of the datasets we are using to conduct this study, we can only show the effectiveness of our method in exploring the solutions space faster. However, previous studies\cite{pakdd} have shown that the SMC DT algorithm can improve the runtime compared to MCMC DT by up to a factor of $343$. We are optimistic that we can achieve the same results, as SMC and EA are inherently parallel algorithms. The main bottleneck of the SMC-EA algorithm is when we evaluate a big number of trees; for example, see the test case of $10$ iterations and $1000$ $T$. We can overcome this problem by the distributed implementation, as we can distribute the trees on many nodes and evaluate the trees concurrently.

\section{Conclusion}
Our study has shown that by combining two novel algorithms, the SMC and EA can tackle major problems on Bayesian Decision Tress and open the space for more research. According to our experimental results, our novel approach based on Sequential Monte Carlo and Evolutionary Algorithms explores the solution space with at least 100 times fewer iterations than a standard probabilistic algorithm, for example, Markov Chain Monte Carlo. As discussed in section~\ref{Results}, we managed to tackle the problem of the naive proposals, and we suggest a method that takes advantage of the communication between the trees. We proposed a sophisticated method to propose new samples based on EA and minimised the burn-in period through SMC.

This study, though, has some limitations. First, as we mentioned, both SMC and EA are inherently parallel algorithms with existing parallel implementations\cite{varsi2020fast,shukla2016parallel, varsi2021log2n}. We plan to extend this study by parallelising the SMC-EA, adding more testing scenarios with larger datasets, and showing improvement in run time.

\bibliography{biblio}

\end{document}